\documentclass[11pt]{article}
\usepackage{coling2020}
\usepackage{times}
\usepackage{url}
\usepackage{latexsym}
\usepackage{amsmath}
\usepackage{todonotes}
\usepackage[utf8]{inputenc}
\usepackage{multirow}
\usepackage{arydshln}
\usepackage{textcomp}
\usepackage{subcaption}
\usepackage{booktabs}

\usepackage{wrapfig,booktabs}
\usepackage{float}
\colingfinalcopy %

\title{Language ID in the Wild: Unexpected Challenges on the Path to a Thousand-Language Web Text Corpus}

\author{Isaac Caswell, Theresa Breiner, Daan van Esch, Ankur Bapna \\
  Google Research, 1600 Amphitheatre Parkway, Mountain View, CA 94043 \\
  {\tt \{icaswell,tbreiner,dvanesch,ankurbpn\}@google.com}}
\date{}

\begin{document}
\maketitle
\begin{abstract}
Large text corpora are increasingly important for a wide variety of Natural Language Processing (NLP) tasks, and automatic language identification (LangID) is a core technology needed to collect such datasets in a multilingual context. LangID is largely treated as solved in the literature, with models reported that achieve over 90\% average F1 on as many as 1,366 languages. We train LangID models on up to 1,629 languages with comparable quality on held-out test sets, but find that human-judged LangID accuracy for web-crawl text corpora created using these models is only around 5\% for many lower-resource languages, suggesting a need for more robust evaluation. Further analysis revealed a variety of error modes, arising from domain mismatch, class imbalance, language similarity, and insufficiently expressive models. We propose two classes of techniques to mitigate these errors: wordlist-based tunable-precision filters (for which we release curated lists in about 500 languages) and transformer-based semi-supervised LangID models, which increase median dataset precision from 5.5\% to 71.2\%. These techniques enable us to create an initial data set covering 100K or more relatively clean sentences in each of 500+ languages, paving the way towards a 1,000-language web text corpus.

\end{abstract}

\blfootnote{
    \hspace{-0.65cm}
    This work is licensed under a Creative Commons Attribution 4.0 International License. License details:    \url{http://creativecommons.org/licenses/by/4.0/}.
}

\section{Introduction}
\label{intro}
Thousands of languages are spoken in our world \cite{Eberhard19-ELW}, but technologies like machine translation (MT) and automatic speech recognition (ASR) are only available in about 100 of them. As internet access becomes increasingly common with the spread of smartphones \cite{Biggs-2017}, bringing technologies that can help lower language and literacy barriers to more languages is ever more important.

Unfortunately, bringing language technologies to more languages is costly, as for many technologies, extending to an additional language has generally required the use of large parallel labeled datasets. For example, ASR systems are usually trained on large sets of audio recordings and transcriptions, while MT systems have historically needed a set of bilingual sentence pairs. Increasingly, small parallel datasets do exist for many languages \cite{Mayer-2014,agic2019jw300,ArtetxeRuder2020,ardila-etal-2020-common}, but those resources were either produced at high cost, or are restricted to narrow domains. Parallel resources, which rarely occur naturally, remain scarce for most languages.

Monolingual text data, which is more commonly produced, is also used in building out language technologies: for example, in training language models, which are used in many applications ranging from next-word prediction in keyboard input software \cite{Ouyang-2017} to ASR and MT \cite{buck2014n}. Historically, though, a monolingual text corpus by itself has not been sufficient to build ASR and MT systems in a new language: at least some parallel data was typically necessary.

Recently, however, significant progress has been made in cross-lingual learning for NLP tasks \cite{Klementiev2012InducingCD,Ammar2016MassivelyMW,lample2019cross,pfeiffer2020madx}: for example, some approaches appear capable of extending machine translation models to new languages with only monolingual data \cite{artetxe2017unsupervised,lample2017unsupervised,Siddhant2020}, and similar findings have been reported for other NLP tasks \cite{hu2020xtreme}. For ASR it is possible to combine a target-language language model with an acoustic model from a phonologically similar language, with no need for parallel datasets of audio recordings and transcriptions \cite{prasad2019}. Such approaches are likely to get even more effective with nearly-universal acoustic models \cite{li2020universal} and more scalable grapheme-to-phoneme modeling approaches \cite{Deri16-GTP,mortensen-etal-2018-epitran,48581,48582,49142,lee-etal-2020-massively}. Even if more work is needed to establish when such approaches will work well \cite{marchisio2020does,ArtetxeRuder2020,wu-dredze-2020-languages}, having useful monolingual text corpora across languages is clearly a prerequisite to exploring such approaches further. Additionally, using techniques such as LaBSE \cite{labse}, parallel corpora can also be constructed from monolingual corpora.

Unfortunately, it has proven challenging to derive highly multilingual text corpora from the web \cite{ArtetxeRuder2020}. One commonly cited reason is that most web content is written in widely-spoken languages like English and Mandarin\footnote{We think existing statistics on the distribution of languages on the web should be taken with a grain of salt, as they were likely gathered using highly imperfect language identification models, as discussed in this paper.}. Still, previous work has shown that the web contains labeled and unlabeled data in  thousands of languages \cite{Scannell-2007,Prasad-18}. Since most web pages do not have any language labels attached, previous efforts to build web text corpora often rely at least in part on crawling selected URLs and top-level domains in each language, or use popular n-gram Language Identification (LangID) models like FastText \cite{FastText} to target a limited number of languages \cite{goldhahn-etal-2012-building,ortizsuarez:hal-02148693}. However, previous work (Section \ref{existing}) has shown that it is possible to build highly accurate LangID systems covering 1,000+ languages.

Thus, aiming to build a 1,000-language web text corpus, we trained a similar large-coverage LangID model, and used it in a large web crawl. However, we found that such LangID systems do not deliver useful results in a real-world web-crawl scenario. To address this, we make the following contributions:

\begin{enumerate}  
    \itemsep0em 
    \item We demonstrate that LangID is much less ``solved" than frequently believed,  and popular n-gram modeling techniques (used for all existing web crawl corpora) have especially serious problems
    \item We categorize common problems LangID models fall prey to (Section \ref{section:cowboy})
    \item We present two improvements over existing approaches: tunable-precision wordlist-filtering and Semi-Supervised Transformer models (Section \ref{section:wblid})
    \item We propose alternative evaluation metrics that better estimate the quality of LangID models from the perspective of web-mining (Section \ref{section:eval}) and perform a deep, 600-language web-crawl (Section \ref{section:crawl})
\end{enumerate}

This work focuses on monolingual corpora, but the problems described also apply to parallel texts, and it is straightforward to extend the improvements described here to parallel data crawling.

\section{LangID Approaches for Web Corpora}
\label{existing}
To create text corpora in as many languages as possible, we needed a broad-coverage, accurate LangID model for our web crawl. We cover existing work and describe our model, built along similar lines.

\subsection{Previous Implementations}
\label{previous}
A rich literature exists on building text corpora from the web: for example, the \textit{Web as Corpus} workshops have focused on the challenges around identifying relevant pages, extracting clean text, content de-duplication, and many other relevant topics \cite{wac-2020-web,jakubicek-etal-2020-current}. We use an internal web crawler, which is equipped with robust text extraction and de-duplication features, and focus on expanding its LangID component.

A comprehensive recent survey on LangID is Jauhiainen et al. \shortcite{survey2018}. Naturally, LangID systems have been applied to web crawls before: Buck et al. \shortcite{buck2014n} published n-gram language models for 175 languages based on Common Crawl data. The Corpora Collection at Leipzig University \cite{goldhahn-etal-2012-building} and the Corpus of Global Language Use \cite{Dunn2020} offer corpora in 252 and 148 languages.  The largest language coverage is probably An Crúbadán, which does not leverage LangID, and found (small amounts of) web data in about 2,000 languages \cite{Scannell-2007}. Our work is probably most similar to OSCAR \cite{ortizsuarez:hal-02148693} and CCNet \cite{ccnet}, which mined Common Crawl data for 166 and 174 language varieties respectively. However, we believe depth of mining and LangID robustness can limit the quality of datasets produced by these projects: a preliminary inspection of the (often small) low-resource language corpora produced by these LangID-based projects discovers the sort of data noise we describe in this paper, which may render them unusable for NLP applications. These Common-Crawl based datasets are also smaller than our final, filtered dataset, which is $\approx$20x larger than CCNet and $\approx$180x larger than OSCAR for shared low-resource languages (see Appendix \ref{appendix:crawl}).

One relevant LangID implementation appearing in the above works is Dunn \shortcite{Dunn2020}, achieving an F1 above 0.95 for 464 languages, and offering a thorough evaluation on different data sources and domains. The only LangID systems with higher coverage that we are aware of are those developed by Brown \shortcite{brown2012finding,brown2013selecting,brown-2014-non}, with the most recent version covering as many as 1,366 language varieties, with accuracy above 99\%. These numbers are impressive, but as we will see, even such high accuracy on test sets will not suffice to derive useful monolingual corpora from a real-world web crawl.

\subsection{Our LangID Implementation}
\label{ours}
The LangID model we built is similar in approach to previously described systems: we use an n-gram based CLD3 model \cite{cld3}, consisting of a single hidden layer feed-forward neural network on bag-of-n-gram features and script-count features, which we trained on an aggregation of proprietary and publicly available text corpora, covering 1,629 language varieties, with an average of 800K tokens per language. Some of the data came from sources with language tags like Wikipedia, while another subset was created using a text elicitation task where we prompted native speakers to write sentences in their language \cite{48745}. For some languages, we also relied on data extracted by Corpus Crawler \cite{corpuscrawler}, a tool which mines text from sites with known in-language content. Using these corpora, we trained several LangID models, on increasingly large sets of languages. As Table \ref{tab:lidf1} demonstrates, even highly multilingual models achieved good F1 scores on held-out test sets.

\begin{wraptable}{r}{6cm}
\begin{center}
\begin{tabular}{|l|l|l|}
\hline
Coverage  & Avg. F1    & Med. F1  \\
\hline
212 lang.   & 96.1\%  & 98.2 \%                     \\
1629 lang.  & 90.4\% & 97.9\%                   \\

\hline
\end{tabular}
\caption{LangID model performance: macro-average F1 and median F1}
\label{tab:lidf1}
\end{center}
\end{wraptable}

We balanced the data to have the same size dataset for each language before training. Since the relatively uncommon languages we are targeting have little web data compared to languages like English, balancing the data makes sense in order to have a high-enough recall model to get whatever scarce data there might be on the web for less common languages. Additionally, practically speaking, weighting training data according to the estimated prevalence of each language on the web at large---for example, with orders of magnitude more English examples than Quechua examples---would likely make model training difficult from a computational and stability perspective. However, it is worth stressing that evaluating a model on balanced data  overestimates the performance of a model on the highly imbalanced web, especially with respect to precision, as we will see in Section \ref{section:mci}.

\section{Failure Modes of LangID Models on Web Text}
\label{section:cowboy}
Despite our LangID models performing well on the held-out test sets, when applied on real-life web data, the models were not as accurate as we had expected. We performed an initial limited crawl with a 648-language model, but some quick evaluations showed that the results were highly noisy, so we performed a full crawl on $\approx$100B documents with a 224-language model to isolate the problems for closer analysis. This model had comparable performance to the models in Table \ref{tab:lidf1}, with median F1 of 96.8 on held-out eval sets. As first-pass filtering, we performed \textbf{document-consistency filtering}: we ran the LangID model on every sentence in each document, and then took the most commonly predicted language as the document language. We only kept sentences where the sentence-level and document-level labels matched. All datasets were also de-duplicated. This approach may have decreased recall on multilingual pages, but it reduced the severe noise problems, and helped reduce disk storage needs.

While we expected some accuracy loss due to the domain mismatch between clean training data and noisy web text \cite{Dunn2020}, even after document-consistency filtering the LangID labels were so noisy that the corpora for the majority of languages in our crawl were unusable for any practical NLP task. Table \ref{fig:noise-classes} presents some representative samples of noise. Beyond various kinds of noise, we also found a high number of unexpected misclassifications, as in the Oromo case in Table \ref{fig:noise-classes}. The following sections detail important classes and sources of noise.

\begin{table}[t!]
\begin{center}
\includegraphics[scale=0.41]{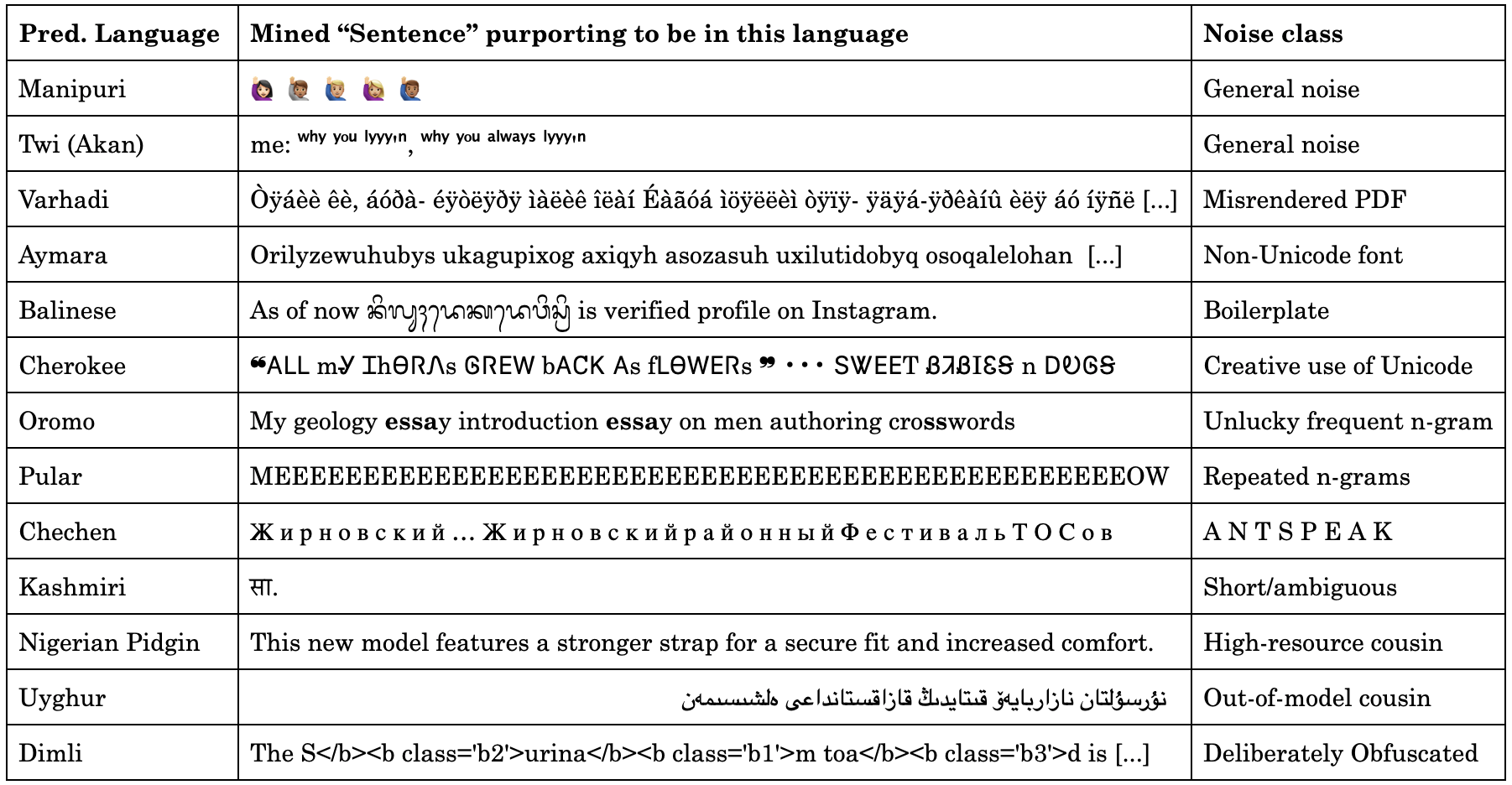}
\caption{Examples of several representative classes of noise in our initial web-crawl corpora.}
\label{fig:noise-classes}
\end{center}
\end{table}

\subsection{Massive Class Imbalances: 99\% Accuracy Is Not Enough}
\label{section:mci}
Precision, unlike recall or false positive rate (\textbf{FPR})\footnote{In the two-class case. FPR depends on the balance of the other classes with respect to each other, but not on the balance of the target class with respect to all other classes. Per-language FPR (e.g. percent of English sentences classified as Nigerian Pidgin) is truly balance-independent.}, is a function of the class balance in a dataset. Measuring precision on a balanced dataset may give misleading impressions about real-world performance. For example, consider a LangID model that has 99\% precision, 99\% recall, and 0.01\% FPR on a particular language on a balanced development set. Imagine however that there are 100 billion pages on the web, of which 10,000 are in the target language: in this scenario, the resulting web-crawled dataset will be mostly out-of-language, containing just under \textit{a tenth of a percent} of sentences in the target language (see calculations in Appendix \ref{appendix:mci})---insufficient for most NLP applications. Yet this assumes a relatively low FPR; for languages with a high FPR with respect to a much more common language, like Nigerian Pidgin with English, the situation is even more dire.

As can be seen from this example, calculations of precision (and by extension, F1) are misleading when applied to real-world data with different class balances than the development set. In the general case, for a classifier with recall $r$ and false positive rate $f$, if we estimate that the language of interest constitutes x\% of the total web text, we get:

\begin{align}
    \textrm{precision}_{crawl} &= \frac{xr}{xr + (1-x)f}
\end{align}

Therefore, any evaluation of LangID models should also report the false positive rate (ideally with respect to major languages on the internet, like English) along with their precision and recall. This class-imbalance effect exacerbates the problems described in the following sections.

\subsection{General Internet Noise and Creativity}
\label{section:noise}
There are many kinds of web noise that are known to cause problems both with LangID and in downstream tasks, such as abbreviations (``g2g", ``hbu"), leetspeak (``n00b"), hashtags (``\#99problems"), or non-standard Unicode encodings (like a \textsc{latin capital letter w} instead of a \textsc{cyrillic capital letter we}). Some of these problems can be handled automatically \cite{Prasad-18,Chua-2018}. However, our efforts in scaling the LangID models in our web crawl to hundreds of languages uncovered greater depths to internet noise, alongside even more creative ways of using text. As a result of the sheer size of the web, any small pathologies of a LangID model are hugely magnified: we observed that our models tend to pick up on particular genres of internet noise for each separate language, resulting in corpora for some languages that mostly showcase a rich array of particular types of oddities.

For example, in our initial crawls, what purported to be the corpus for Varhadi picked up large amounts of badly-encoded PDFs; Aymara and Turkmen were made up mostly of misrendered non-Unicode text; Dimli had mostly invalid HTML; Dogri offered a rich array of Zalgo-like ornamentation; Fula was awash in URLs; Ilocano caught vast amounts of garbled Javascript; and Zhuang captured German sentences involving the Unicode \textsc{soft hyphen} character. In each of these cases, sadly the majority of the crawled corpus actually consisted of the class of noise that the LangID classifier decided to assign to these languages---unfortunately drowning out any in-language sentences in the corpora.

In another interesting twist, one might expect that languages which are written in scripts that are not used for any other language would have clean corpora, as the unique connection between the script and the language means that any LangID model gets 100\% F1 on development sets. However, this underestimates the creativity of the internet: the Cherokee syllabary, for example, contains characters that look similar to Latin characters, which are consequently repurposed to give words in other languages an aesthetic effect (see example in Table \ref{fig:noise-classes}), while other scripts, such as Balinese, are used commonly for purely decorative purposes alongside content in entirely unrelated languages. Some script-unique languages like Divehi do yield high-precision corpora right from the get-go, but they are the lucky few.

\subsection{Artifacts from Character N-gram Modeling}
\label{section:ngram_artifacts}

Many error modes seem to be direct consequences of n-gram count based models, and are also common in public corpora crawled using n-gram models like FastText \cite{FastText}---Appendix \ref{appendix:oscar} explores these phenomena in the OSCAR \cite{ortizsuarez:hal-02148693} corpus. Here are a few important classes of pathologies we discovered; see Table \ref{fig:noise-classes} for examples of each, and Appendix \ref{appendix:noise-stats} for frequency statistics:

\begin{enumerate}
    \item{\textbf{Unlucky overlap of frequent n-grams with high-prevalence languages:}} Token frequencies in natural text follow a power law distribution \cite{zipf}, so that the most common n-grams in a language will be present in a majority of all of its sentences. If one of these common n-grams happens to occur in a sentence in a different language, LangID models can over-trigger. We observed this with Oromo, where 50\% of the crawled dataset was actually English sentences containing the word ``essay" \textit{at least three times}, misleading the model due to high counts for the n-grams ``essa", ``ess", ``sa", ``a", ``e", ``s", and ``y", all of which are top Oromo n-grams (see Appendix Table \ref{tab:essay}).
    \item{\textbf{Repeated n-graaaaaaaaams:}} By repeating an n-gram sequence an arbitrary amount, which is rare in clean training text but common on the internet, the class probability of a language may be ramped up, even if the language is clearly wrong---cf. adversarial examples \cite{goodfellow2015explaining}.
    \item{\textbf{A N T S P E A K :}} A surprisingly common internet phenomenon is to find text with space-separated characters, \mbox{l i k e  t h i s} \cite{ants}. Standard n-gram models--or even SentencePiece models \cite{kudo2018sentencepiece}--can't handle this without special-casing. This affects about one to two languages per major script: we found that most of our ``Chechen" data was actually \mbox{R u s s i a n}, most of our ``Lambadi" \mbox{T e l u g u} , our ``Santali" \mbox{B e n g a l i}, and some of our ``Sepedi" \mbox{E n g l i s h}.
\end{enumerate}

\subsection{Languages with High-Prevalence Cousins}
Languages with High-Prevalence Cousins is a specific, quite common case of the Class Imbalance problem, which requires somewhat different techniques to mitigate (see Section \ref{section:wblid}). Crawling the web for a low-resource language (\textit{``target language"}) that is closely related to a language that is highly prevalent on the internet (\textit{``distractor language"}) can yield a dataset consisting mostly of the distractor language. A particularly salient example is Nigerian Pidgin (i.e. Naija, `pcm') and English (`en'), which are similar enough (see Appendix Table \ref{tab:naija} for examples) that typical LangID models will have high false positive rates between the two. Because of the prevalence of English on the internet, along with this high degree of confusability, building a high-precision web-crawled text corpus for languages like Nigerian Pidgin is exceedingly difficult.

\subsection{Languages with Out-of-Model Cousins}
A variant on the above are languages that are not supported by the LangID model, which interfere with related languages that are supported. For example, a majority of our Uyghur crawl was actually Kazakh and Kyrgyz in the Arabic script; our model had been trained to recognize Kazakh and Kyrgyz, but only in the Cyrillic alphabet. Table \ref{fig:noise-classes} gives an example Kazakh sentence that was labeled as Uyghur.

\subsection{Unrepresentative Training Data}
 Sometimes training data may be \textit{too clean} to be accurate on out-of-domain, noisy web data; yet other times it may be too noisy, too homogeneous, or contain systematic biases.  For example, for some languages, training data (especially data sourced from Wikipedia) had high quantities of special characters and templated data (esp. from censuses). Templated data may be harmful for n-gram models, by skewing the token distributions away from that of normal text, though there is some evidence that neural models may be less affected by token distributions than by latent structure \cite{papadimitriou2020pretraining}. Other training data may also have issues; for instance, in our elicited Chechen data, the \textsc{cyrillic letter palochka} (not found on many keyboards) was represented with the ASCII digit ``1". Our model therefore may not handle Chechen text containing the correct code point, or other substitutes, very well.

\section{Improving LangID Precision on Web Text}
\label{section:wblid}
Monolingual web-text corpora afflicted by the issues described in Section \ref{section:cowboy} will likely prove unusable for practical purposes. We report on two distinct approaches we found helpful in improving precision.

\subsection{Tunable-precision Filtering with Curated Wordlists}
We experimented with token-based filtering techniques, which are simple to implement and fast to perform on large corpora. Since the LangID models in our crawl operated on character n-grams, token-based approaches may have complementary behavior and can side-step particular failure modes. For instance, since a sentence with the word ``essay" likely contains mostly non-Oromo words, the havoc caused by the n-gram ``essa" described in Section \ref{section:ngram_artifacts} is neatly sidestepped by checking against a curated list of known Oromo words. Such filtering approaches have the added benefit of {\it tunable precision}, allowing us to adjust the cleanliness of our corpora depending on the noise tolerance of downstream tasks.

\subsubsection*{Percent-Threshold filtering}
The simplest approach to token-based filtering is to remove any sentence where less than $x$\% of its tokens appear in a clean list of known words for the language, such as one would find in a standard dictionary. We used in-house lists with a median of $\approx$15K words per language, which were obtained through frequency sorting followed by human curation. The one parameter for filtering---the percentage of in-vocabulary words---provides a simple, interpretable way to tune for precision/recall. We call this method \textbf{Percent-Threshold Wordlist Filtering}.

\subsubsection*{TF-IDF based filtering}
Percent-Threshold Wordlist Filtering is effective for a majority of the problems we saw, where the text is nonsense or in an entirely different language, but it will \textit{not} help where the mislabeled text is in a similar language, as in Nigerian Pidgin (`pcm'), which has very high lexical overlap with English (`en')---meaning that such filtering will still retain most English sentences, and fail to increase precision. This problem will occur with any language that has high lexical overlap with a major language. Where there is extensive borrowing of loanwords, the languages may even be unrelated, as for Chuvash and Russian.

Some words, however, are highly effective language markers: for example, ``wetin" is common in Nigerian Pidgin, but does not occur in English. We therefore propose to keep any sentence that has at least one word from a small list of common tokens that are \textit{distinctive} to that particular language, and are not shared with its more prevalent cousins. We call this \textbf{Disjunctive Wordlist Filtering}.

First, we perform \textsc{tf-idf}, where each ``document" is our LangID training set. However, this suffers one crucial flaw: the idf formulation of \textsc{tf-idf} weights each document equally, so a word will be equally penalized if it occurs in English or in K'iche'. For practical purposes, we care mainly about filtering out common distractor-language text on the internet, so we only want to penalize those languages.

This motivates a simple variant on \textsc{tf-idf} which we call \textsc{tf-iif}, or \textbf{Term Frequency-Inverse Internet Frequency}. This measure is the ratio of the frequency of a token in our per-language corpus (\textsc{tf}) with the frequency of that token across the entire internet (\textsc{iif}), which we approximate from a sample of 7 million randomly selected web sentences. In practice we find that performance improves slightly when accounting for both \textsc{idf} and \textsc{iif}, yielding the \textsc{tf-idf-iif} score. Formally, for a token $t$ in a language $l$, with a frequency function $f(term, corpus)$ and language-specific corpora $D_l$:

\begin{equation}
\textrm{tf-idf-iif}_{t,l} = \textrm{tf}_{t,l} * \textrm{idf}_{t,l}  *  \textrm{iif}_{t} = f ( t, D_l ) \log \left( \frac{1}{\sum_{l' \ne l} 1\{t \in D_l'\}} \right) \frac{1}{f ( t, internet )}
\end{equation}

With a ranked \textsc{tf-idf-iif} list for each language, we then pick the top N words for each language such that we have at least $r$\% recall on our dev sets. While it is tempting to choose the same $r$ for all languages (e.g. 95\%), different languages can behave quite differently with such filters, with small changes in recall sometimes leading to large changes in precision. We had best results by choosing $r \in [0.75, 1.0]$, and then determining the ideal precision-recall trade-off on a per-language basis. With this paper, we publicly release \textsc{tf-idf-iif} wordlists we used, covering the top 100 tokens for each of about 500 languages\footnote{https://github.com/google-research-datasets/TF-IDF-IIF-top100-wordlists}.

\subsection{Semi-Supervised LangID}
\label{section:sslid}

A separate approach from filtering is to improve our original LangID model. Utilizing large unsupervised text corpora to improve the quality of neural networks has become increasingly important in NLP \cite{devlin2018bert,wang2018glue}. Following this line of work, we use the noisy data crawled with our n-gram LangID model to improve the quality of our LangID system by leveraging self-supervised approaches, yielding a Semi-Supervised LangID system (\textbf{SS-LID}).

\begin{wraptable}{r}{9.1cm}
\begin{center}
\begin{tabular}{|l|r|r|r|r|}
\hline
model  & \multicolumn{1}{|l|}{F1} &  \multicolumn{1}{|l|}{prec.}   &  \multicolumn{1}{|l|}{rec.} & \multicolumn{1}{|l|}{FPR}           \\
\hline

NG-LID$_{212}$	& 96.05	& 	94.93	& 	97.64	& 	0.01079 \\
\hdashline				
XF-LID$_{212}$	& 97.51	& 97.26	& 97.82 &  0.00849 \\
\multicolumn{1}{|r|}{$\Delta \varepsilon$} 	& 36.9\% 	& 46.0\% &	7.4\% 	& 21.4\% \\
SS-LID$_{212}$	&  98.03	&  97.61	&  98.55	& 	0.00683 \\
\multicolumn{1}{|r|}{$\Delta \varepsilon$} 		&  50.2\% 	& 	52.9\%	&  38.4\%	&  36.7\% \\
SS-LID$_{624}$		& 97.52	& 	97.86	& 	97.45	& 	0.00610 \\
\multicolumn{1}{|r|}{$\Delta \varepsilon$} 	& 	37.3\%		& 57.8\%	& -8.3\%	& 43.5\% \\

\hline
\end{tabular}
\caption{Performance of n-gram LangID model, Transformer LangID model (XF-LID) and Semi-supervised models (SS-LID) trained on either 212 or 624 languages. Scores are averaged over the shared 212 languages.}
\label{tab:sslid}
\end{center}
\end{wraptable}

Specifically, following the text-to-text self-supervised approach outlined in Raffel et al. \shortcite{raffel2019exploring}, we train a Transformer Big model \cite{vaswani2017attention} by sampling equally from the crawled data from 212 languages. We co-train this self-supervised task with the LangID task in a text-to-text setting, with the hope of improving the quality of LangID on noisy open-domain web text. To reduce the confounding effect of using a higher capacity transformer, we train a baseline transformer on just the LangID task.

We evaluate these SS-LID models and compare against the n-gram based LangID model in Table \ref{tab:sslid}. In addition to F1, precision, and recall, we report FPR, whose importance we discussed in Section \ref{section:mci}. All values are macro-averaged over the shared 212 languages. To distinguish between apparently well-performing models we also report the relative error reduction with respect to the n-gram model, which for an error metric $\varepsilon$ we define as  $\Delta \varepsilon = \frac{\varepsilon_{b} - \varepsilon_{t}}{\varepsilon_{b}}$, where $\varepsilon_{b}$ is the baseline model error and $\varepsilon_{t}$ the test model error.

We see that the Transformer LangID model outperforms the n-gram model by a large margin, especially on precision and FPR. The SS-LID models improve further upon this model, notably with a ~40\% reduction in FPR. It is worth noting that these improvements are on the \textbf{clean} eval set, despite the additional training objective being on the noisy web crawl. We suspect the improvements are even greater on web-type data, which is partially validated by the evaluation on web-text in Section~\ref{section:eval}.

\section{Evaluating LangID Filtering Methods on Web-Text}
\label{section:eval}

\subsection{Evaluation Methodology: Principles and Suggestions}
Ideally, LangID models would be evaluated on a large, noisy test set, representative of real-life web data. Since such sets do not currently exist, we recommend having human annotators evaluate crawled corpora to ensure quality meets the threshold for downstream use (which will vary per application). For automatic metrics, we suggest focusing on false positive rate and recall rather than precision and recall, and comparing models using relative error reduction to amplify differences between apparently highly-performant models, as we did above in Section \ref{section:sslid}.

\subsection{Evaluating our Systems}
We asked human annotators to evaluate LangID quality for our web-crawled text in a subset of the languages. First, we filtered the web crawl with several methods. We then randomly sampled 100-1,000 sentences from each of these filtered data sets, and asked annotators (who were fluent speakers, or who spoke a closely related language) to indicate whether each sentence was in the target language.

\begin{table}[]
\begin{center}
\begin{tabular}{|l|rr|rr|rr|rr|rr|}
\hline
 & \multicolumn{2}{c|}{Unfiltered}   & 	\multicolumn{2}{c|}{Threshold}  & 	\multicolumn{2}{c|}{Disjunctive} & 	\multicolumn{2}{c|}{SS-LID$_{624}$}  \\ 
Language     & \multicolumn{1}{c}{$p$} & \multicolumn{1}{c|}{$r$} & \multicolumn{1}{c}{$p$} & \multicolumn{1}{c|}{$r$} & \multicolumn{1}{c}{$p$} & \multicolumn{1}{c|}{$r$} & \multicolumn{1}{c}{$p$} & \multicolumn{1}{c|}{$r$}\\
\hline					
\textbf{Aymara} & 2.1 & 100 &  86.2 & 98.3 &  76.4 & 92.9 &  \textbf{94.6 }& 99.3 \\   
\textbf{Bhojpuri}   &	4.0 & 100& 	3.0 & 100 & 	4.0 & 93.0 & 	\textbf{83.0} & 98.5 \\ 
\textbf{Chechen} &  24.0 & 100 & 84.0 & 99.9 & 49.0 & 91.8 & \textbf{98.0} & 99.9 \\ 
\textbf{Cherokee} & 16.0 &  100 & 95.0 & 100 & \textbf{97.0} & 90.6 &  47.0 & 100 \\ 
\textbf{Chuvash} &  5.0 & 100 &  3.0 &  43.2 & \textbf{22.0} & 93.5 &  5.0 & 99.5 \\  
\textbf{Divehi} & 98.8 & 100 &  98.6 & 99.0 &  \textbf{99.1} & 91.4 &  98.9 & 99.6 \\ 
\textbf{Guarani} &  4.0 & 100 &  12.0 &  99.0 & \textbf{44.0} & 92.1 &  23.0 & 98.8 \\  
\textbf{Oromo} &  5.0 &  100 & 78.0 & 98.0 &  \textbf{80.0} & 91.8 &  33.0 & 99.5 \\
\textbf{Surjapuri} & 31.3 & 100 &  45.9 & 97.1 &  \textbf{61.0} & 88.2 &  60.3 & 95.6 \\
\textbf{Swiss German} & 2.0 & 100 &  2.0 &  98.7 & 2.0 & 92.1 &  \textbf{16.0} & 95.5 \\
\textbf{Tamazight} &  6.0 & 100 &  \textbf{42.0} & 98.8 &  35.0 & 91.3 &  \textbf{42.0} & 99.8 \\ 
\textbf{Twi (Akan)} & 49.0 & 100 & \textbf{83.0} &  100 & \textbf{83.0} & 92.4 &  82.0 & 93.5 \\  
\textbf{Zhuang} & 1.0 & 100 &  \textbf{59.0} & 85.9 &  3.0 & 92.6 &  15.0 & 98.3 \\    
\hdashline
\textbf{Median} & 5.5 & 100 &  52.5 & 98.5 & 47.5 & 92.0 & \textbf{71.2} & 98.7 \\  
\hline
\end{tabular}
\caption{Comparison of filtering approaches for a few languages: percent-threshold fitering ($x=20\%$), disjunctive \textsc{tf-idf-iif} filtering ($r=90\%$), and filtering with a Semi-supervised LangID model. We report 1.~human-judged LangID precision over the crawl (percent of in-language sentences), and 2.~recall of this method on our held-out eval sets. Best precision is bolded. Full table in Appendix.}
\label{tab:human-eval-table}
\end{center}
\end{table}

Table \ref{tab:human-eval-table} presents the results of this evaluation for a selection of languages (full results on seventeen languages in Appendix Table \ref{tab:human-eval-table-appendix}). For each language, we show the precision of the method from the human annotations, and the recall of the same filter on our clean dev sets. For the percent-threshold filtering we evaluated a threshold of 20\%, and for the disjunctive wordlist filtering we used the top N \textsc{tf-idf-iif} words per language such that the recall on our held-out eval set was at least 90\%.

We see that the initial datasets were extremely noisy, with a median value of 5\% of sentences being in-language. The filtering methods drastically increased the percentage of correctly LangID'd sentences, with values of up to 99\% in-language, while maintaining high recall. However, the best filtering method varies widely by language. The neural SS-LID model has the highest precision for Bhojpuri and Swiss German, both of which also suffer most from the High-Prevalence-Cousin issue among these languages. However, it does much more poorly than wordlist-based approaches on Oromo and Cherokee. In the latter case, we found that SS-LID was unable to discard English sentences written in Cherokee syllabics.

It is worth re-emphasizing that the thresholds in Table \ref{tab:human-eval-table} were chosen somewhat arbitrarily for the purpose of illustration. Since precision is tunable in the word-based approaches, precision can be increased further, though at growing cost to recall---a trade-off to make depending on downstream noise tolerance.

For Guinea-Bissau Creole, which has both a High-Prevalence Cousin (Portuguese) and an Out-of-Model Cousin (Papiamentu), none of our filtering methods were effective (see Appendix). Swiss German, in the same situation, barely scraped by. Future work should investigate additional techniques for such cases---although the most effective solution may be as simple as using a hand-curated \textsc{tf-idf-iif} list, which looked promising in preliminary experiments in Nigerian Pidgin.

\section{Web-crawled Dataset and Comparison with other Public Datasets}
\label{section:crawl}
Using the above methods\footnote{Our process is also summarized in Appendix \ref{appendix:recipe} for those interested in replicating.}, we performed a deep crawl of the web (touching $>$100B webpages) with a 600-language LangID model. Using percent-threshold filtering\footnote{In this case, we used larger wordlists than those used for the analysis above, in order to stress recall.} we made a recall-focused dataset, then post-filter with a SS-LID model for high precision, yielding a larger, cleaner set than is found in similar corpora. More details and comparisons to public corpora (OSCAR, CCNet) are in Appendices \ref{appendix:oscar} and \ref{appendix:crawl}.

\section{Future Work}
\label{section:future}
Our approach yielded usable monolingual text corpora in $\approx$600 languages. Internal user experience research suggests the web may now contain at least some amount of monolingual text in thousands of languages, so we plan to scale up with more multilingual LangID models, like our 1,629-language model.

Truly covering the linguistic richness of the web will also need crawling approaches to be fine-tuned further. Text for some languages may only be found in PDF files \cite{bustamante-oncevay-zariquiey:2020:LREC}, and some scripts are commonly represented in non-Unicode fonts---such as Kruti Dev for Devanagari, requiring separate detection for conversion into Unicode-encoded Devanagari \cite{singhfont}. Applying OCR may also help handle non-Unicode text, and can uncover textual content within images. And many languages that are not officially written in the Latin alphabet have informal transliterated orthographies \cite{roark-etal-2020-processing}; our models can identify the most common ones, but we could cover more.

Finally, our work focused on a web crawl, but many new internet users primarily use their language online on social media platforms and in chat messages \cite{DLDP,48745}. Other work has looked at applying LangID to social media \cite{jaech2016hierarchical,blodgett2017a,vo2019language}. Our techniques should help improve LangID accuracy in this challenging domain, too.

\section{Conclusion}
Language Identification (LangID) is by no means a solved problem, and n-gram models are much worse than popularly believed. We trained LangID models covering up to 1,629 languages, but found that even seemingly high-quality models ($>95$ F1) were nearly unusable in practice for low-resource languages. We described and analyzed several major issues encountered in applying LangID to a real-life web crawl. These practical problems included large amounts of noise, much of which appears to be natural language and can't be easily filtered out; insufficient expressiveness of n-gram models; issues with related languages; and a massive class imbalance problem, meaning that even 99\% F1 can be insufficient.

To solve these issues, we developed two major improvements to our LangID system: tunable-precision filtering methods (for which we release wordlists in about 500 languages) and semi-supervised neural models. These allowed us to create usable monolingual text corpora across hundreds of languages based on our deep web crawl, with much more and cleaner data per language than previously published approaches. Such corpora hold great promise for bringing technologies like MT and ASR to more languages, and we believe it should be possible to use the approaches we outlined to create monolingual corpora in many more languages, which should help extend language technology even further.

\section*{Acknowledgements}
We would like to thank Diana Akrong, Alex Rudnick, Mikhail Donolin, Maxim Krikun, Hakim Sidahmed, and Landis Baker for help with human evaluations of the LangID models, as well as Vera Axelrod, Jason Riesa, and Wolfgang Macherey for useful advice and reviews. We also want to specifically thank Onome Ofoman for her consultation and advice about Nigerian Pidgin.

\bibliographystyle{coling}
\bibliography{coling2020}
\clearpage

\appendix

\clearpage

\section{Complete human evaluation results}
\label{appendix:complete_human_eval}
A more complete version of Table \ref{tab:human-eval-table} is given here in Table \ref{tab:human-eval-table-appendix}, containing the full set of seventeen languages we evaluated. The only additional information it shows over Table \ref{tab:human-eval-table} is the percentage of the web-crawl each method filters out, for more context into how these methods will behave in practice. (Keep in mind that, while the precision and \% filtered rows are measured on the noisy web crawl, the recall is measured on the held-out eval set.)

\begin{table}[H]
\begin{center}
\begin{tabular}{|l|rrr|rrr|rrr|rrr|rrr|}
\hline
 & \multicolumn{3}{c|}{Unfiltered}   & 	\multicolumn{3}{c|}{Threshold}  & 	\multicolumn{3}{c|}{Disjunctive} & 	\multicolumn{3}{c|}{SS-LID}  \\ 
Language     & \multicolumn{1}{c}{$p$} & \multicolumn{1}{c}{$r$} & \multicolumn{1}{c|}{$filt.$} & \multicolumn{1}{c}{$p$} & \multicolumn{1}{c}{$r$} & \multicolumn{1}{c|}{$filt.$} & \multicolumn{1}{c}{$p$} & \multicolumn{1}{c}{$r$} & \multicolumn{1}{c|}{$filt.$} & \multicolumn{1}{c}{$p$} & \multicolumn{1}{c}{$r$} & \multicolumn{1}{c|}{$filt.$} \\
\hline					
\textbf{Ahirani*} & 49.1 & 100 & 0.0 &  38.2 & 100.0 & 27.2  & 46.0 & 90.7 & 49.8 &  \textbf{96.4} & 98.6 & 86.6 \\
\textbf{Aymara} & 2.1 & 100 & 0.0 &  86.2 & 98.3 & 98.3  & 76.4 & 92.9 & 98.2 &  \textbf{94.6 }& 99.3 & 98.1 \\
\textbf{Bashkir*} & 33.1 & 100 & 0.0 &  84.9 & 95.2 & 62.1 & \textbf{ 91.5} & 91.9 & 67.7 &  89.7 & 99.4 & 61.0 \\
\textbf{Bhojpuri}   &   4.0 & 100& 0.0 &      3.0 & 100 & 28.3     & 4.0 & 93.0 & 57.6 &        \textbf{83.0} & 98.5 & 97.0 \\
\textbf{Chechen} &  24.0 & 100 & 0.0 & 84.0 & 99.9 & 73.9 & 49.0 & 91.8 & 69.1 & \textbf{98.0} & 99.9 & 78.7 \\
\textbf{Cherokee} & 16.0 &  100 & 0.0 & 95.0 & 100 & 86.9 & \textbf{97.0} & 90.6 & 87.7 &  47.0 & 100 & 68.6 \\
\textbf{Chuvash} &  5.0 & 100 & 0.0 &  3.0 &  43.2 & 56.3 & \textbf{22.0} & 93.5 & 89.5 &  5.0 & 99.5 & 54.8 \\
\textbf{Divehi} & 98.8 & 100 & 0.0 &  98.6 & 99.0 & 2.7  & \textbf{99.1} & 91.4 & 27.2 &  98.9 & 99.6 & 0.0 \\
\textbf{Guarani} &  4.0 & 100 & 0.0 &  12.0 &  99.0 & 77.4 & \textbf{44.0} & 92.1 & 93.5 &  23.0 & 98.8 & 85.4 \\
\textbf{G.B. Creole* † } & 0.0 &  100 & 0.0 & 0.0&      100 & 18.5 & 0.0&    93.6 & 36.1 & 0.0&   92.9 & 76.3 \\
\textbf{Kinyarwanda*} & 37.3 & 100 & 0.0 &  79.6 & 93.0 & 58.1  & 88.1 & 91.9 & 62.0 &  \textbf{90.9} & 98.8 & 60.8 \\
\textbf{Oromo} &  5.0 &  100 & 0.0 & 78.0 & 98.0 & 99.0  & \textbf{80.0} & 91.8 & 99.0 &  33.0 & 99.5 & 97.4 \\
\textbf{Surjapuri} & 31.3 & 100 & 0.0 &  45.9 & 97.1 & 34.7  & \textbf{61.0} & 88.2 & 51.8 &  60.3 & 95.6 & 77.9 \\
\textbf{Swiss German} & 2.0 & 100 & 0.0 &  2.0 &  98.7 & 70.6 & 2.0 & 92.1 & 43.3 &  \textbf{16.0} & 95.5 & 88.5 \\
\textbf{Tamazight} &  6.0 & 100 & 0.0 &  \textbf{42.0} & 98.8 & 88.2  & 35.0 & 91.3 & 78.6 &  \textbf{42.0} & 99.8 & 88.5 \\
\textbf{Twi (Akan)} & 49.0 & 100 & 0.0 & \textbf{83.0} &  100 & 50.0 & \textbf{83.0} & 92.4 & 42.5 &  82.0 & 93.5 & 38.9 \\
\textbf{Zhuang} & 1.0 & 100 & 0.0 &  \textbf{59.0} & 85.9 & 98.9  & 3.0 & 92.6 & 84.6 &  15.0 & 98.3 & 88.2 \\
\hdashline
\textbf{Median} & 5.5 & 100 & 0.0 &  52.5 & 98.5 & 60.1 & 47.5 & 92.0 & 64.9 & \textbf{71.2} & 98.7 & 82.6  \\ 
\hline
\end{tabular}

\caption{More complete comparison of different filtering approaches for different languages. For each example language, we report 1. the precision of the crawl (percent of in-language sentences), as judged by human raters over a sample of 100 sentences per filtering method, 2. the recall of this method on our held-out eval sets, and 3. the percentage of the crawl removed by this filtering method. * Starred languages were omitted from the table in the main paper. † G.B. = Guinea-Bissau}
\label{tab:human-eval-table-appendix}
\end{center}
\end{table}

\section{Massive Class Imbalance: Worked Example}
\label{appendix:mci}
This section shows the methodology for the example in Section \ref{section:mci}, where we examine by way of example a LangID model with 99\% precision, 99\% recall, and 0.01\% FPR for a given language. If we approximate that there are 100 billion pages on the web, of which 10,000 are in a language we are seeking, we can analyze the precision of the web crawl using the quantities of True Positives (TP), True Negatives (TN), False Negatives (FN), and False Positives (FP). For the dataset resulting from the web crawl, we can therefore say that $TN + FP  \approx 100B - 100k \approx 100B$, and  $TP + FN \approx 100k$. One can now calculate $p_{crawl}$, the precision on the resulting crawl of the web:

\begin{align}
TP &= \frac{TP}{TP + FN}(TP + FN) = r * (TP + FN) = 0.99*10k = 9.9k \\
FP &= \frac{FP}{TN + FP} (TN + FP) = fpr * (TN + FP) = 0.0001*100B = 10M \\
p_{crawl} &= \frac{TP}{TP + FP} = \frac{9.9k}{9.9k + 10M} \approx 0.1\%
\end{align}

\section{Statistics on languages most affected by different types of noise}
\label{appendix:noise-stats}
Many of the types of noise mentioned in Section \ref{section:noise} are hard to quantify without significant extra work. For instance, it would require building special classifiers for misrendered PDFs, non-Unicode fonts, creative use of Unicode, and so on---and it may need a stronger classifier than an n-gram classifier, since after all these are mistakes of an n-gram classifier. Issues like out-of model-cousins are even trickier, probably requiring human ratings. However, some types of noise can be quantified using approximations like the following:

\begin{itemize}
    \itemsep0em 
    \item { \bf A N T S P E A K :} regex match with \verb|  /[^ ] [^ ] [^ ] [^ ] [^ ]/ |
    \item { \bf n-graaaaams:} regex match with \verb|/((.)\2\2\2\2)/| up to \verb| /((.....)\2\2\2\2)/|  
    \item { \bf HTML: } regex match with \verb|/<[a-z/]*>/|
    \item { \bf http: } regex match with \verb|/http/|
    \item { \bf Title Case: } $>5$ successive tokens such that \verb|x[0].isupper() and x[1:].islower()|
    \item { \bf essay: } (special for Oromo) regex match with \verb|/[Ee]ssay/|
    \item { \bf misrendered PDF:} contains bigrams along the lines of \verb|{åí,íè,ñò}| etc. or \verb|{^j,j^,^J}| etc. (basically, we created a very simple bigram classifier on known misrendered PDFs)
\end{itemize}

\begin{table}[H]
\begin{center}
\begin{tabular}{|l|l|l|}
\hline
Language (Script) & Phenomenon   & Percent of crawl  \\
\hline
Lambadi (Telu.)  & A N T S P E A K  & 72.1\%                     \\
Santali (Beng.) & A N T S P E A K & 58.2\%                   \\
Bodo (Beng.) & n-graaaaams & 50.9\%                   \\
Pular  & n-graaaaams & 26.3\%                   \\
Avar  & HTML & 64.2\%                   \\
Dimli  & HTML & 93.1\%                   \\
Fula  & http & 44.5\%                   \\
Magahi  & http & 23.6\%                   \\
Nigerian Fulfulde  & Title Case & 64.5\%                   \\
Balinese  & Title Case & 63.1\%                   \\
Oromo  & essay & 64.4\%                   \\
Varhadi  & misrendered PDF & 90.8\%                   \\
Yucateco  & misrendered PDF & 74.7\%                   \\
\hline
\end{tabular}
\caption{Quantification of the incidence of a few noise phenomena, along with their most affected languages in our web-crawl.}
\label{tab:quantified-phenomena}
\end{center}
\end{table}

\section {Details on the web-mined datasets}
\label{appendix:crawl}
As described in Section \ref{section:crawl}, the dataset we mined has two versions, one focused on recall (called {\tt recall} in the table), and one focusing on precision (called {\tt sslid(recall)} in the table). Table \ref{tab:crawl} compares these two datasets with public benchmarks. 

Since the purpose of this crawl was to focus on low-resource languages, we mined a smaller portion of the internet for the $\sim$100 highest-resource languages, and did not do any filtering on these languages. For this reason, in addition to the stats on the entire dataset, we report the stats on the dataset omitting the highest-resource 100 languages, to give a fairer approximation of the size of datasets for truly low-resource languages. We also report stats on the languages among those that are shared between the three datasets, again omitting the $\sim$100 highest resource languages.

Please note that these datasets are hard to compare to public benchmarks, as they crawl a wider swath of the internet, and are much more highly multilingual. Therefore, the comparison with public data sources in this table should not be interpreted as giving information about the nature of the filtering methods described in this paper.

\begin{table}[H]
\begin{center}
\begin{tabular}{|l||l|l|l||l|l|l||l|l|l||}
\hline
 metric &  \multicolumn{3}{c||}{N Languages} & \multicolumn{3}{c||}{N Sentences} & \multicolumn{3}{c||}{Median Dataset size} \\ 
\hdashline
 subset & all & 100+ & shared & all & 100+ & shared & all & 100+ & shared \\
 \hline
 recall & 600 & 500 & 59 & 36B & 3200M & 3200M & 2100K & 970K & 12000K \\
 SS-LID(recall) & 600 & 500 & 59 & 2.8B &  600M & 740M & 200K & 100K & 1400K \\
 CCNet & 174 & 74 & 59 & 70B & 4.4M & 22M & 930K & 6K & 78K \\
 OSCAR & 166 & 66 & 59 & 20B & 0.5M & 5.4M & 200K & 1K & 8K \\
 \hline
\end{tabular}
\caption{Comparison between the two versions of our dataset and the public datasets CCNet and OSCAR. Although the statistics look similar on the full dataset, we see that the public datasets are heavily skewed towards higher-resource languages. When excluding the 100 highest-resource languages (``100+"), or looking only at shared low-resource languages (``shared"), we see that the public datasets have 20x to 200x less data than our crawl was able to identify.}
\label{tab:crawl}
\end{center}
\end{table}

\section{Comparison with OSCAR Corpus}
\label{appendix:oscar}
While the analyses in the main paper focused on evaluating the quality of the data we crawled, publicly available datasets have similar issues. This section briefly analyzes the OSCAR corpus \cite{ortizsuarez:hal-02148693}, which, although an excellent resource for many languages, has lower-quality content for some languages. All analyses are performed on the deduplicated OSCAR corpus, which is cleaner.

\begin{wraptable}{r}{8.2cm}
\begin{center}
\begin{tabular}{|l|l|l|}
\hline
Language & Phenomenon   & \% of crawl  \\
\hline
Central Bicol  & A N T S P E A K  & 100.0\%                     \\
Neapolitan & A N T S P E A K & 100.0\%                   \\
Emilian-Romagnol & A N T S P E A K & 55.8\%                   \\
Somali & n-graaaaams & 88.1\%                   \\
Cantonese  & n-graaaaams & 57.1\%                   \\
Asturian  & n-graaaaams & 53.0\%                   \\
\hline
\end{tabular}
\caption{Most-affected languages in the OSCAR corpus for two common error modes of n-gram models}
\label{tab:oscar-ngram}
\end{center}
\end{wraptable}

Please note that it is hard to compare OSCAR directly with our dataset. One notable confound is that the two datasets are drawing from different portions of the web. Another confound is the degree of multilinguality and the subset of languages chosen (this paper tends to focus on longer-tail languages than OSCAR). A further large confound is that OSCAR uses the FastText LangID model \cite{FastText}, which does not upsample training data, and therefore will tend to have lower recall and higher precision.  

\begin{figure*}
\centering
\begin{subfigure}{.5\textwidth}
  \centering
  \includegraphics[width=.9\linewidth]{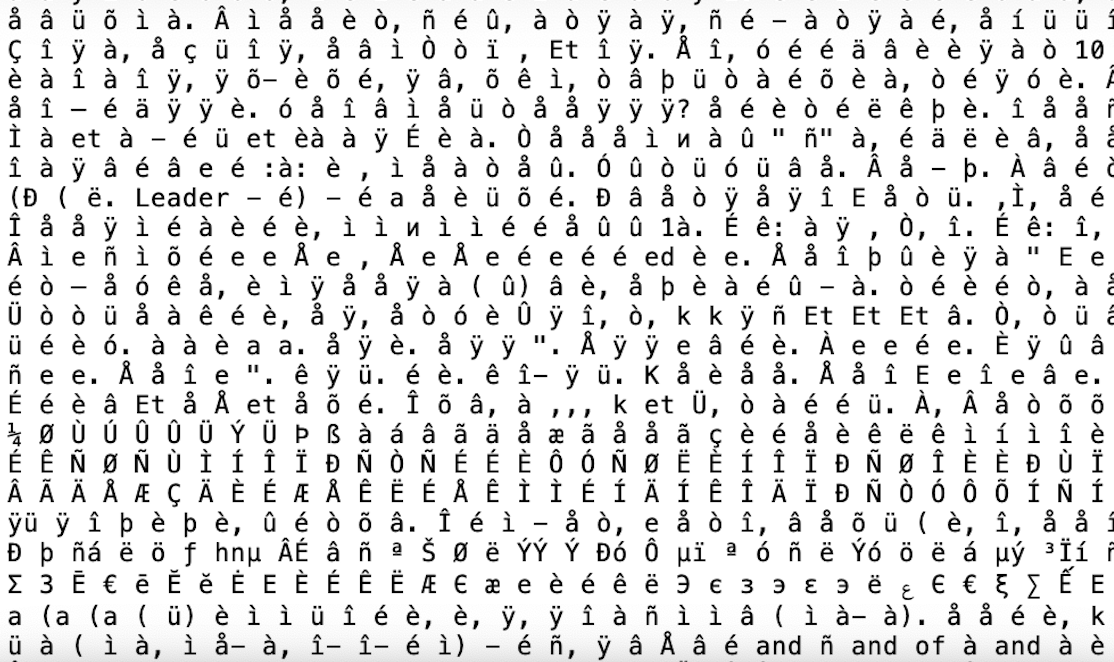}
  \caption{``Neopolitan" (actually A N T S P E A K - like content)}
  \label{fig:sub1}
\end{subfigure}%
\begin{subfigure}{.5\textwidth}
  \centering
  \includegraphics[width=.9\linewidth]{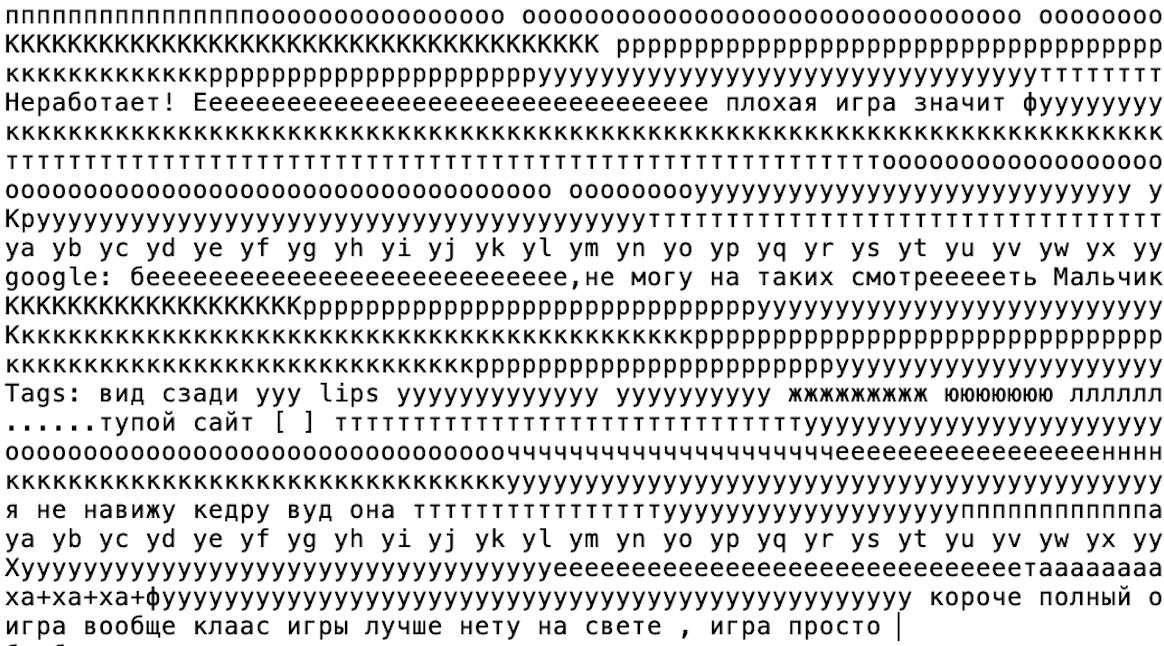}
  \caption{``Somali" (actually repeated ngraaaaaaaaams)}
  \label{fig:sub2}
\end{subfigure}
\caption{Representative samples from OSCAR corpora affected by two n-gram LangID error modes}
\label{fig:oscar}
\end{figure*}

Applying the heuristic analyses from Section \ref{appendix:noise-stats}, we see that repeated ngram and A~N~T S~P~E~A~K issues are also very common in the OSCAR corpus (the other phenomena from Table \ref{tab:quantified-phenomena}, however, were mostly absent). Table \ref{tab:oscar-ngram} reports the three most affected languages per phenomenon, and Figure \ref{fig:oscar} shows a representative sample of two of these corpora. In both these cases, the dataset consisted only of such noise, and had no in-language content.

To further analyze the cleanliness of the OSCAR corpus, we performed a similar analysis as in Section \ref{section:eval}, to determine the percentage of each dataset that was in-language. Table \ref{tab:oscar-wblid} summarizes these findings, along with the percentage of the corpus remaining after percent-threshold filtering with our wordlists. We only look at the thirty lowest-resource languages in the corpus. We find that the percent in-language varies widely by language, ranging from 0\% to 100\%. However, many of the corpora have relatively high precision, with the average precision being just over 89\%. At the same time, this accords with a low average recall, with the median dataset size being only 37 sentences. It is interesting to note that wordlist-filtering corresponds quite well with human-judged precision, with Pearson's R of 87.3\%.

\begin{table}[H]
\begin{center}
\begin{tabular}{|l|r|r|r|}
\hline
Language & 	precision & 	wordlist-match & 	N \\
\hline
Central Bikol & 	0\% & 	0\% & 	1 \\
Chavacano & 	0\% & 	0\% & 	1 \\
Dimli & 	100\% & 	100\% & 	1 \\
Pampanga & 	100\% & 	100\% & 	2 \\
Bavarian & 	25\% & 	25\% & 	4 \\
Erzya & 	100\% & 	100\% & 	5 \\
Mirandese & 	57.1\% & 	N/A & 	7 \\
Yue Chinese & 	14.3\% & 	57.1\% & 	7 \\
Northern Frisian & 	0\% & 	0\% & 	9 \\
Haitian & 	30\% & 	30\% & 	10 \\
Interlingue & 	15.4\% & 	0N/A & 	11 \\
Sicilian & 	100\% & 	100\% & 	17 \\
Tuvinian & 	96.2\% & 	96.2\% & 	26 \\
Maithili & 	89.7\% & 	89.7\% & 	29 \\
Russia Buriat & 	100\% & 	97.3\% & 	37 \\
Lower Sorbian & 	97.6\% & 	95.1\% & 	41 \\
Somali & 	0\% & 	0\% & 	42 \\
Romansh & 	100\% & 	100\% & 	47 \\
Nahuatl languages & 	100\% & 	30\% & 	60 \\
Neapolitan & 	0\% & 	0\% & 	61 \\
Yoruba & 	100\% & 	100\% & 	64 \\
Guarani & 	81.5\% & 	81.5\% & 	81 \\
Venetian & 	91.4\% & 	91.4\% & 	81 \\
Cornish & 	89.2\% & 	91.6\% & 	83 \\
Wu Chinese & 	0\% & 	68.6\% & 	86 \\
Bihari & 	89.4\% & 	95.2\% & 	104 \\
Emilian-Romagnol & 	43.3\% & 	42.3\% & 	104 \\
Northern Luri & 	99.1\% & 	87.6\% & 	113 \\
Limburgan & 	99.2\% & 	96.9\% & 	128 \\
Minangkabau & 	56.1\% & 	57.2\% & 	180 \\
\hdashline			
Median & 	89.3\% & 	89.7\% & 	39 \\
\hline
\end{tabular}
\caption{The 30 lowest-resource languages in OSCAR, and 1. their human-judged percent in-language (i.e. precision); 2. the percentage remaining after applying percent-threshold wordlist filtering; and 3. total number of sentences in the (deduplicated) corpus. Languages for which we lacked wordlists are marked with ``N/A".}
\label{tab:oscar-wblid}
\end{center}
\end{table}

\section{Notes on Curated Wordlist Approaches}
For languages written in unsegmented scripts (where spaces are not used in between words; for example, Mandarin), leveraging the curated wordlists during the filtering techniques is not as straightforward. When given a sentence to check for valid words, we would first need to run a segmentation model in order to split the sentence into words, but segmentation models need to be trained on specific languages and do not usually support lower-resource languages. To handle languages written in such writing systems, we included all valid characters in the language as part of the wordlist, so that we could fall back to character-level checks for any sentences written in these scripts. This means that any somewhat reasonable language data using the same script will be kept, even if it is a different language.

\section{Wordlist-based Language ID}
\label{appendix:pure_wblid}
For languages with little or no training sentence-level data, even an n-gram LangID model is not practical to train. We therefore additionally explored pure wordlist-based models: specifically, we experimented with a Word-Based LangID system (\textbf{WB-LID}), which assigns a LangID label to the sentence by simply counting how many known words appear in the sentence for each possible language and predicting the language with the highest counts, with extra weight granted to ``unique words" that appear only in a single language's wordlist. The simple architecture of WB-LID does not compare to an n-gram LangID model for most languages (Table \ref{tab:wblid-results}), and we decided not to pursue using the outputs of WBLID as a filter in this work, but this approach seems stable and scalable to more languages, and may be worth exploring in the future as a LangID system for languages where no sentence data can be found to train an n-gram model.

\begin{table}[H]
\begin{center}
\begin{tabular}{|l|l|l|}
\hline
LangID system Comparisons (median F1) & 493 Languages & 590 Languages                      \\
\hline
n-gram LangID & 97\% & 96\%                      \\
Word-Based LangID & 75\% & 76\%                     \\

\hline
\end{tabular}
\caption{Performance of the n-gram LangID system vs Word-Based LangID system on development sets. For the dev sets shown in this comparison, we only include languages for which we had both sentence data to train the n-gram model and known wordlists to train the WBLID. We remove any known words from our WBLID system that do not appear in the sentence data used to train the n-gram model. The n-gram model is trained on all sentence data for the supported languages.}
\label{tab:wblid-results}
\end{center}
\end{table}

\section{Illustration of the High-Prevalence-Cousin problem}
\label{appendix:pcm}

Although the issue of highly similar varieties is very common and may be familiar to speakers of most languages in the world, English-speaking researchers may be less familiar with it, since close relatives of English do not generally receive a lot of attention in the literature. As an illustration, Table \ref{tab:naija} gives some examples of Nigerian Pidgin and the English translations. It is clear that a simple classifier might have trouble distinguishing them, especially for more technical sentences.

\begin{table}[H]
\begin{center}
\begin{tabular}{|l|l|}
\hline
\textbf{Nigerian Pidgin }                                                   & \multicolumn{1}{l|}{\textbf{English}}                                                 \\ \hline
abeg, you fit help me?    & please, can you help me?  \\ 
\hline
no dey buy wetin we no need    & \multicolumn{1}{l|}{don't go buying what we don't need} \\ \hline
He don accuse her family say dem inflate di value & \multicolumn{1}{l|}{He has accused her family of inflating the value} \\ \hline
Born August 28, 1991 & Born August 28, 1991 \\ \hline
Structured, goal-oriented education & Structured, goal-oriented education \\ \hline
\end{tabular}
\caption{Examples of Nigerian Pidgin versus English. It is very hard to mine datasets of Nigerian Pidgin from the web, because it is close enough to English that Language ID models and frequent-wordlist filtering methods will pick up a lot of English. In the informal register, like the first few examples, they are more distinguishable, but in the formal, written register they can appear identical.}
\label{tab:naija}
\end{center}
\end{table}

\section{Oromo: A Case Study in Unfortunate N-gram Overlap}
\label{appendix:oromo}
As alluded to in Section \ref{section:ngram_artifacts},  Oromo has the peculiar error mode that our n-gram model massively over-triggers with English, despite the two languages bearing little to no resemblance to each other, as a result of the frequent 4-gram ``essa". Table \ref{tab:essay} illustrates this further, showing the most common n-grams in true Oromo, in natural English, and in the web-crawl that claimed to be Oromo.

\begin{table}[H]
\begin{center}
\begin{tabular}{|ll|ll|ll|}
\hline
Oromo        &  & ``Oromo"      &  & English         &  \\ 
 LID        & idx. &  crawl     & idx. &  LID        & idx. \\ 
\hline
atti    & 0     & ssay  & $>$1000 & your  & $>$1000 \\
anii    & 1     & essa  & 4     & have  & $>$1000 \\
akka    & 2     & tion  & $>$1000 & with  & $>$1000 \\
eess    & 3     & writ  & $>$1000 & what  & $>$1000 \\
essa    & 4     & atio  & $>$1000 & here  & $>$1000 \\
jedh    & 5     & mple  & $>$1000 & ther  & $>$1000 \\
isaa    & 6     & ment  & $>$1000 & tion  & $>$1000 \\
oota    & 7     & ampl  & $>$1000 & want  & 930 \\
kees    & 8     & tive  & $>$1000 & like  & $>$1000 \\
itti    & 9     & ting  & $>$1000 & thin  & $>$1000 \\
\hline
\end{tabular}
\caption{Top 10 most common 4-grams in a) Oromo LangID training data, b) the ``Oromo" crawl of the web, and c) English LangID training data. Each 4-gram is presented with its index among the top 1,000 most common Oromo 4-grams. We can understand from the n-gram list that the ``Oromo" crawl is majority English, overtriggering because of the 4-gram ``essa", from the English word ``essay". In fact, 50\% of sentences in the ``Oromo" crawl contain the word ``essay" at least three times! The other common n-grams in this Table from the ``Oromo" crawl are epiphenomenal, reflecting only English words that tend to occur in English sentences about essays. }
\label{tab:essay}
\end{center}
\end{table}

\section{Correlation of filtering precision with relevant variables}
\label{appendix:correl}
When do some filtering methods work better than others? We do not have enough data points to make strong statements (N=17), but there are some trends that may be worth commenting on here. In Table \ref{tab:correl}, we look at the correlation of the precision of unfiltered data and the three proposed filtering methods, and how they correlate with 1) the size of the crawled dataset, and 2) the dialectical relatedness to common languages online. We hypothesize that variable (1) is a combination of variable (2) with non-linguistic noise artifacts, so looking at these two variables can give us an idea of which methods are better at general noise filtering (from train-data pathologies, etc.) and distinguishing related languages.

Unfortunately the ``dialectical relatedness to common languages online" is hard to quantify. As a rough approximation, we introduce four heuristic ``confusability classes":

\begin{enumerate}
    \itemsep0em 
    \item { \bf Class 1: } No obviously confusable languages
    \item { \bf Class 2: } Confusable low-resource languages or slightly confusable high-resource language
    \item { \bf Class 3: } Medium-confusable high-resource language
    \item { \bf Class 4: } Very confusable high-resource language
\end{enumerate}

To perform the regression we assign these classes to the values \{1, 2, 3, 4\}. Per-language assignments are given in Table \ref{tab:confusability_score}.

Based on the numbers in Table \ref{tab:correl}, it looks like both wordlist filtering methods perform similarly, and the SS-LID method is noticeably better when languages are more confusable, and possibly slightly worse when there are larger datasets (signalling more confusion with non-linguistic or out-of-domain noise).

\begin{table}[H]
\begin{center}
\begin{tabular}{|l|rr|}
\hline
& log(n. segs) &  confusion rank \\  
\hline
unfiltered & -0.71  & -0.16 \\  
threshold & -0.26 & -0.75 \\  
disjunctive & -0.21  & -0.46 \\  
SS-LID & -0.55 &  0.15 \\
\hline
\end{tabular}
\caption{Pearson correlation of the precision of three filtering methods (and unfiltered data) with two relevant variables. Number of segments (i.e. number of sentences in the ``unfiltered" dataset) is passed through a log transform first, since the size of the unfiltered datasets follows a log distribution. For an explanation of the ``confusion rank", please see the appendix section \ref{appendix:correl} and Table \ref{tab:confusability_score}. }
\label{tab:correl}
\end{center}
\end{table}

\begin{table}[H]
\begin{center}
\begin{tabular}{|l|l|l|}
\hline
 Language & confusion class & Notes / relevant related languages \\  
\hline
Divehi & class \#1 & script unique \\  
Zhuang & class \#1 & pretty unique orthography \\  
Cherokee & class \#2 & some confusion with ``Cherokee English" etc. \\  
Guarani & class \#2 & some lexical overlap with Spanish \\  
Tamazight & class \#2 & some lexical overlap with ar-Latn and Tamasheq \\  
Twi(Akan) & class \#2 & some lexical overlap with Ewe, Ga, etc. \\  
Kinyarwanda & class \#2 & high lexical overlap with Rundi etc.  \\  
Aymara & class \#2 & some lexical overlap with Spanish \\  
Oromo & class \#2 & some lexical overlap with Gedeo, Hamer, Somali etc. \\  
Bashkir & class \#3 & medium lexical overlap with Russian \\  
Ahirani & class \#3 & medium lexical overlap with Hindi \\  
Chechen & class \#3 & medium lexical overlap with Russian \\  
Surjapuri & class \#3 & medium lexical overlap with Hindi \\  
Chuvash & class \#4 & high lexical overlap with Russian* \\  
Bhojpuri & class \#4 & high lexical overlap with Hindi \\  
Guinea-Bissau Creole & class \#4 & high lexical overlap with Portuguese \\  
Swiss German & class \#4 & high lexical overlap with German \\  
\hline
\end{tabular}
\caption{Heuristic judgement of ``confusability" for use in the regression in Table \ref{tab:correl}. Please note that this is not a rigorous quantification of these languages and may contain mistakes. For explanations of the ``classes", please see text. * Note that Chuvash is considered ``high" overlap because of polluted training data.}
\label{tab:confusability_score}
\end{center}
\end{table}

\section{Complete Recipe}
\label{appendix:recipe}
This section is simply a concise description of the steps we took to create our dataset, in the form of suggestions for someone interested in creating a similar dataset.
\begin{enumerate}
    \item { \bf Train LangID model}
    \begin{enumerate}
      \item \textbf{Balance the data first} in order to have higher recall. The distribution of languages in training data may not be representative of the distribution of languages on the web. Temperature sampling \cite{arivazhagan2019massively} may also be a good alternative, in order to decrease overtriggering somewhat.
      \item If it is computationally feasible to apply a more complex model at inference time, a Transformer-based LangID model (especially co-trained with a self-supervised objective on in-domain text) will have better performance, even if the held-out scores seem only slightly better.
      \item \textbf{Evaluate cannily:} use out-of-domain held-out sets if possible, and pay special attention to the relative reduction in false-positive rate. A model with FPR of 0.1 is much different than one with FPR of 0.01---don't give up once you reach 95\% F1.
    \end{enumerate}
   \item \textbf{Curate wordlists.} If the publicly released wordlists don't suit one's purposes, one could take e.g. the 200 most frequent tokens from the train set, removing words that are also in highly-prevalent languages if desired, like English, Portuguese, Spanish, Russian, German, Chinese, and Hindi. One can skip this step if the Transformer LangID model is good enough, but it will still be useful for tuning the precision of the final datasets, and will still improve for several languages (e.g. in our situation, it was necessary to catch English written in Cherokee script).
  \item \textbf{Perform the web crawl.}  Document-consistency filtering is highly recommended (only output sentences whose sentence-level ID matches the majority sentence-level ID on the page).
  \item \textbf{Deduplicate the web-crawled data} and filter with wordlists to reach a desired precision.
   \item \textbf{Look at samples of every language in the dataset!} Even quickly eyeballing the dataset can reveal serious problems. Also consider quickly checking that all the language codes are plausible: for instance, is the { \tt als} data a mix of Tosk Albanian (ISO639-3 code { \tt als}) and Swiss German (which Wikipedia stores under the code {\tt als})? Or are there some macrolanguage codes in the dataset that cover a superset of other already-covered languages, like Norwegian Bokmal {\tt nb}, Norwegian Nynorsk {\tt nn}, and the macrolanguage code Norwegian {\tt no}?
\end{enumerate}

\end{document}